Paper ID 253

# Using SlowFast Networks for Near-Miss Incident Analysis in Dashcam Videos

**Yucheng Zhang [1, *], Koichi Emura [1], Eiji Watanabe [2, 3]**

1. R&D Division, Panasonic Automotive Systems Co., Ltd., Yokohama City, Japan
2. Laboratory of Neurophysiology, National Institute for Basic Biology, Okazaki City, Japan
3. Department of Basic Biology, The Graduate University for Advanced Studies (SOKENDAI), Hayama City, Japan

[1, *]zhang.yucheng@jp.panasonic.com, [1]emura.koichi@jp.panasonic.com, [2, 3]eiji@nibb.ac.jp

**Abstract**

This paper classifies near-miss traffic videos using the SlowFast deep neural network that mimics the characteristics of the slow and fast visual information processed by two different streams from the M (Magnocellular) and P (Parvocellular) cells of the human brain. The approach significantly improves the accuracy of the traffic near-miss video analysis and presents insights into human visual perception in traffic scenarios. Moreover, it contributes to traffic safety enhancements and provides novel perspectives on the potential cognitive errors in traffic accidents.

**Introduction**

With the rapid development of urbanization and motorization comes the increasingly notable social issue of road traffic safety encountered by the current society. Traffic accidents not only cause significant human loss, but it also places a heavy burden on the socioeconomy. Therefore, research and implementation of effective traffic safety measures are greatly important. In recent years, technological progress, such as event data recorders (EDR) and advanced driver assistance systems (ADAS) has played a crucial role in road safety enhancement. EDR and ADAS collect and analyze driving behavior data to determine how a vehicle behaves before and after a collision, enhance the vehicle performance, and detect immediate dangers, such as suddenly crossing pedestrians, alerting drivers, and activating automatic braking, which significantly reduce traffic accident rates. However, existing technologies are still limited when it comes to the processing of complex traffic scenes and the prediction of potential dangers (e.g., traffic near-miss incidents).

Traffic near-miss incidents are situations that do not result in accidents but have potential risks. They are an important study area in traffic safety research because they typically act as accident indicators and provide valuable data that prevent future incidents. The Society of Automotive Engineers of Japan built a traffic near-miss incident dataset [1] that assists researchers and companies in analyzing traffic near-miss accidents and reducing traffic accidents. This dataset has now been transferred to the Tokyo University of Agriculture and Technology (TUAT) Drive Recorder Data Center [2]. Yamamoto et al. [3] developed several deep learning-based classification methods that can identify near-miss and traffic accidents. However, their approaches often require a combination of other sensor data, such as speed and GPS data, which are not commonly available. Consequently, relying solely on dashcam videos is insufficient.

In this study, we propose a new method for addressing the abovementioned issues. We use the SlowFast [4] network to analyze the traffic Near-Miss Incident dataset. This network is an advanced deep learning model for analyzing video data and simulating the way the human brain processes visual information. It can simultaneously capture rapid changes (fast) and slow-moving information (slow) in videos.

The study contribution is a new deep learning-based method for classifying traffic near-miss incidents. This study reports the significant advantages of processing complex spatiotemporal data when compared to existing technologies (e.g., method in [3]). We analyze the results of the SlowFast network by exploring the relationship between human visual cognition in traffic environments and demonstrate how this method can contribute to improving traffic safety. Our research provides new perspectives and approaches for the prevention and identification of future traffic accidents.

The Related Work section provides an overview of the existing methods in the traffic safety field, particularly the EDR and ADAS application, and discusses their limitations in identifying traffic near-miss incidents. We introduce research on neural network application in traffic safety, especially for the SlowFast network. The Method section describes the dataset used in this work, SlowFast network customization, and training process. The Experiments and Results section first quantitatively compares our SlowFast network results to those in obtained in [3], and then demonstrates the accuracy of identifying the traffic near-miss incidents using precision, recall, and F1 scores. Next, we conduct a qualitative analysis comparing regions within the video frames classified as traffic near-misses visualized using Grad-CAM [5], with areas of attention in the same frames estimated using DeepGazeIIE [6] and explore their connection with human vision. Subsequently, we discuss the limitations of the method and directions for future research. Finally, the Conclusion section summarizes the main findings and study contributions and proposes future research expectations.

**Related Work**

The recent years have witnessed an increase in the application of deep learning technologies in the traffic safety field. In particular, neural networks yield an outstanding performance in the image analysis area by classifying complex traffic scenes. Deep learning shows special advantages when processing large amounts of unstructured data. For a dynamic video that includes movement and changes, such as that captured by dashcams, deep learning can effectively identify and predict potentially near-miss incident situations. As an advanced neural network, the SlowFast network is especially effective in video data analysis [4]. It combines sensitivity to fast movements by understanding contextual information at slow speeds and mimics the human visual processing method, making it particularly suitable for the traffic scene analysis.

Visual cognition plays an important role in understanding and explaining traffic scenes. Watanabe et al. [7]

successfully reproduced the illusory motion by using the PredNet deep neural network. Inspired by Watanabe et al. [7], Emura et al. [8] and Kato et al. [9] used the PredNet model, which pre-trained by [7], to generate visual illusions to explain the human perception principles in attempt to reduce the occurrence of cognitive errors in traffic accidents. This influenced our use of the SlowFast network to analyze the traffic near-miss incidents herein. The magnocellular (M) cells in the human visual neural system are believed to be sensitive to fast movements, while the parvocellular (P) ones focus more on the color, texture, and complex details [10][11][12]. Inspired by the application of similar mechanisms in traffic safety research using deep learning models that mimic human vision, we attempt herein the development of more accurate and effective traffic near-miss incident prevention technologies by simulating these biological characteristics.

The novelty of this study is the usage of the SlowFast network, a neural network combining sensitivity to dynamic scenes and the ability to understand complex contexts to analyze traffic near-miss incidents. Compared to existing techniques (e.g., method combining sensor data and video analysis in [3]), our approach demonstrates a higher accuracy when processing video data only. Our method uses dashcam videos, which are more commonly available and reachable data. Compared to methods requiring additional sensor data, our approach is more flexible and scalable.

This study employs a unique approach of understanding and simulating the human visual system method of processing complex traffic environment information. The SlowFast network allows our method to capture rapidly changing events (e.g., emergency braking and sudden changes in direction) and understand the contextual environment related to these events (e.g., road conditions and detailed traffic situations). The method of integrating high and low-speed visual streams offers a new perspective in the identification and analysis of traffic near-miss incidents. Our study is innovative in its methodology. We focus on the network's performance representation (e.g., classification accuracy) and explore how the network is connected to mechanisms of human visual perception. This multidisciplinary approach provides new insights in the traffic safety field, particularly in understanding and predicting traffic near-miss incident situations.

In summary, the main study contribution is the proposal of a new traffic near-miss classification method based on a video analysis that demonstrates an advanced and innovative understanding of complex traffic environments. Unlike existing methods, our method shows enhanced capabilities in processing spatiotemporal data and offers identification efficiency and accuracy improvements. It also provides new perspectives and tools for future traffic safety research and practice.

**Method**

In this study, we used the Near-Miss Incident dataset [2] provided by the Tokyo University of Agriculture and Technology. The dataset had videos collected from taxi dashcams across various regions of Japan, covering urban and rural environments and a range of seasons and weather conditions. The scenes are diverse and show scenarios from everyday driving to potential near-miss events. The dataset was created using triggers like emergency braking, emergency steering, and sudden acceleration.

The dataset features [2] are as follows: 1) scale: it contains over 200,000 independent traffic events each represented by a 15-second video clip (Figure 1); note that we only used 287 independent data samples. This specific number was chosen for a rapid initial analysis, focusing on an average selection of the five most frequent traffic accident scenes in Japan: crossing collision, right turn, left turn, pedestrian crossing, and rear-end collision [13]; 2) diversity: the data cover various traffic scenes, including urban traffic rush, rural roads, nighttime driving, and driving under adverse weather conditions; and 3) annotation: each traffic near-miss video clip has detailed annotations made by experts, which include information on the event type, time and location of the near-miss occurrence, and relevant details about the movements of nearby vehicles and pedestrians.

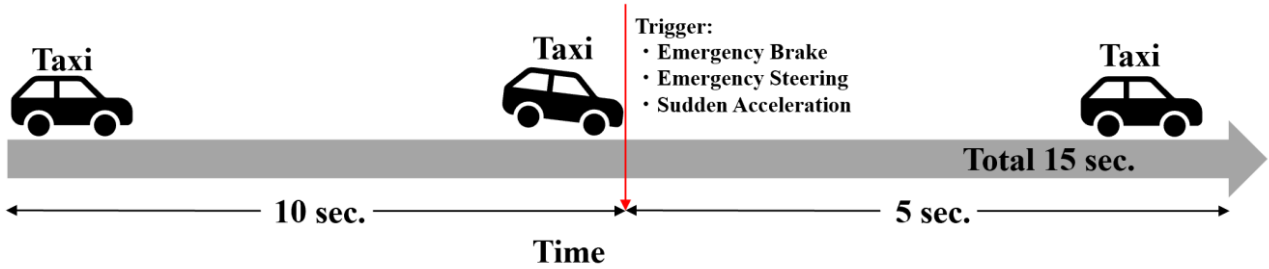

Figure 1 - **Near-miss incident data sample**

In data preprocessing, 287 time-series video data were divided into segments representing the traffic near-miss incident and safe driving durations. Within the original 15-second video clip, the moment occurring at the 10th second of each clip represented the traffic near-miss event. Different time intervals were set to distinguish between traffic near-miss events and safe driving. T represents the total time of the original video clip. $T_{near-miss}$ is the time duration classified as a traffic near-miss event. $T_{safe-driving}$ is the time duration classified as safe driving. Each 15-second video clip is denoted as T = 15 s, with the interval from 0 to less than 5 seconds as $T_{safe-driving}$ = [0 s, 5 s), and the interval from 5 to 10 seconds as $T_{near-miss}$ = [5 s, 10 s].

The 287 video data samples were divided into different groups and split into a 6:2:2 ratio for training, validation, and testing to avoid data leakage. This resulted in a data distribution of 172:57:58 (train:validation:test). Data augmentation techniques (e.g., Scale Jitter) were used during training to improve the model generalization capacity.

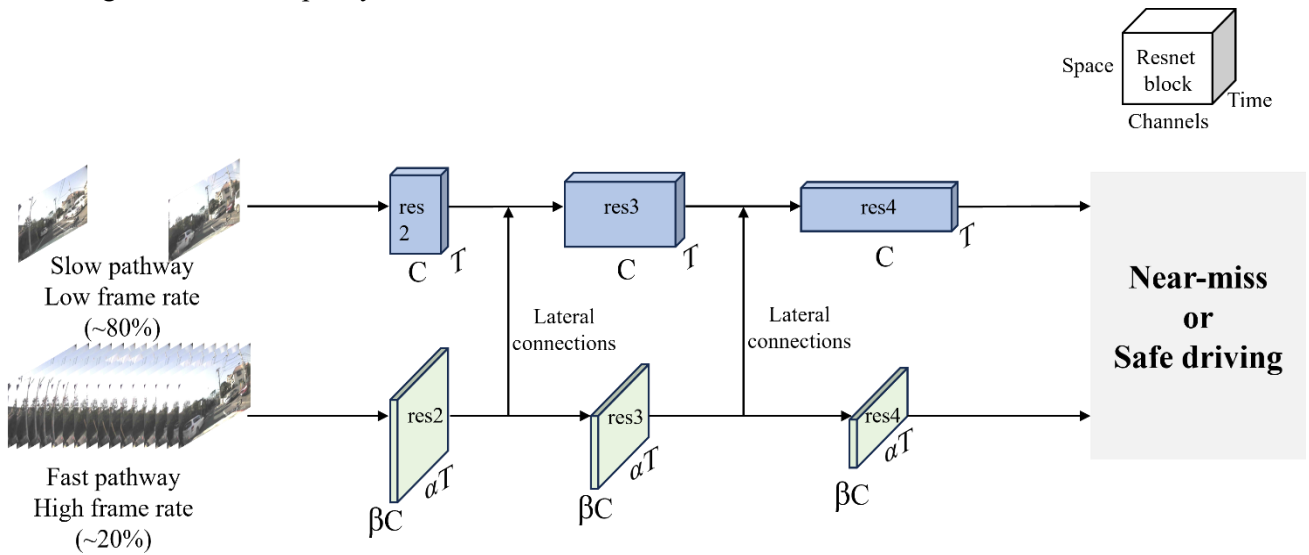

Figure 2 - **SlowFast model**

Using SlowFast Networks for Near-Miss Incident Analysis in Dashcam VideosAs for the model construction, we used the SlowFast [4] network to analyze the traffic near-miss dataset. Figure 2 shows the SlowFast architecture. For a video clip, the Slow pathway operated at a low frame rate (e.g., 2 frame) and covered approximately 80% of the total computation, while the Fast pathway operated at a high frame rate (e.g., 8 frame) and occupied approximately 20% of the total computation. This ratio was aligned with the fact that M cells composed approximately 20%, while P cells were approximately 80% [4][10][11][12].

We chose SlowFast over other models because it was parallel to the human visual system, wherein the M cells were sensitive to rapid movements, while the P cells were more focused on the color, texture, and complex details [10][11][12]. The SlowFast network mimicked this biological characteristic with its dual-stream structure. Analyzing the traffic near-miss incident scenes allowed not only for rapid responses to urgent situations, such as sudden braking or emergency steering, but also for an effective understanding of the complex contextual environment related to the events, including road conditions, traffic flow, and visible traffic signs.

Compared to the original SlowFast network [4], we made several changes based on the feature of the Near-Miss Incident dataset. In the three-dimensional convolutional neural network (CNN) block, we chose ResNet101 [14] to increase the network depth, enabling it to better capture the complex traffic scenes and small-scale dynamic changes. We also integrated a non-local [15] module to enhance the network capacity and capture distance-dependent relationships between distant objects within video frames, particularly improving the classification of significant events and objects.

In the parameter setting, we set α to 4 and β to 8. We used a warmup strategy to ensure training stability and efficiency and enhance the final model performance. The warmup period continued for 34 epochs during which the learning rate gradually increased from a beginning value of 0.01 to 0.1. Following this, we used a cosine annealing strategy to gradually decrease the learning rate from its beginning value to reach the optimal results. Under this strategy, the learning rate steadily decreased until the maximum epoch number of 196 was reached. Subsequently, it began to increase again. For the network optimization, the learning rate was calculated following Equation (1), where $Lr_t$ is the learning rate at the t-th training cycle; $Lr_{min}$ is the minimum learning rate; $Lr_{max}$ is the maximum learning rate; $T_{current}$ is the current number of training cycles; and $T_{max}$ is the maximum number of training cycles within a complete cosine cycle. The maximum number of training cycles was set at 196. The loss function used was cross-entropy. The optimization method was stochastic gradient descent. The dropout rate was 0.5. The momentum was 0.9. The weight decay was set at $1e^{-4}$.

$$Lr_t = Lr_{min} + \frac{1}{2}(Lr_{max} - Lr_{min})(1 + \cos(\frac{T_{current}}{T_{max}}\pi)) \qquad (1)$$

During the training process, a high-performance computer equipped with an NVIDIA GPU 3080Ti was used to ensure sufficient computational resources and training efficiency. The PyTorch deep learning framework, specifically PyTorch version 1.13.1 with CUDA 11.7 support, was utilized in combination with CUDA and cuDNN acceleration libraries to achieve efficient network training and optimization.

**Experiments and Results**

In the experiment, the model was trained from scratch without using pre-trained weights. It underwent training for 196 epochs during which the dataset fully passed through the network for 196 times. After the training completion, the model achieved a final accuracy of 66.67% on the test dataset. Figure 3 depicts a graph of the model Top1 error, wherein the horizontal axis represents the number of iterations during the training process. Each iteration refers to the data batch processing and model updating. The model gradually learned from the training data as the iterations progressed. The vertical axis represents the Top1 error rate, which gradually decreased as the iterations progressed.

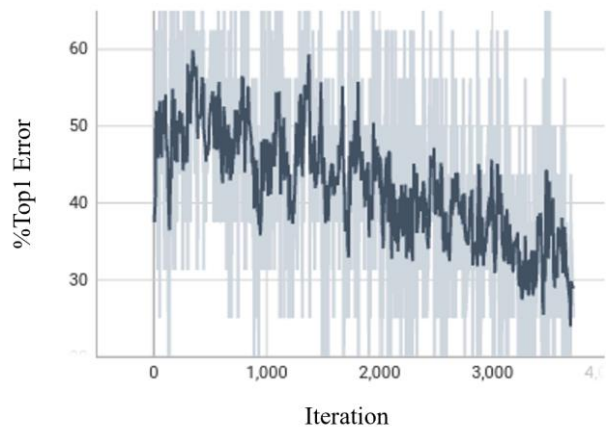

**Figure 3 - Top1 error during training**

Figure 4 depicts a graph of the loss during the model training, where the vertical axis loss shows the differences between the model predictions and the actual labels. The goal during training was to minimize this loss value. The loss value gradually decreased as the iterations were being processed, indicating that the model was learning from the data. Overall, both the loss and the Top1 error decreased. Figures 3 and 4 illustrate a large wave motion smoothed by a filter. This wave motion could be attributed to the dataset containing challenging problems to learn because traffic near-miss incidents can be hard to distinguish, even for humans.

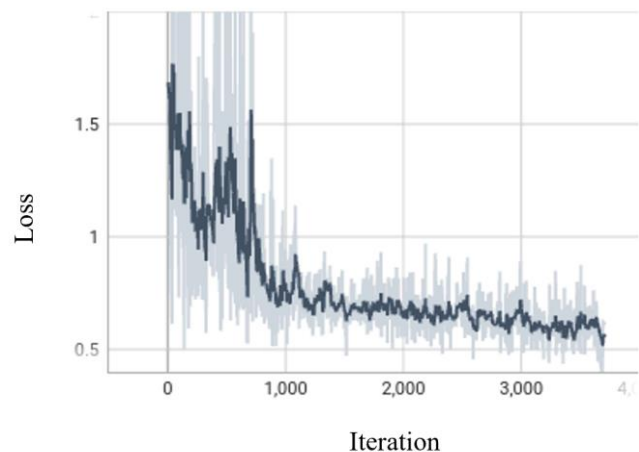

**Figure 4 - Loss during training**

Table 1 presents the confusion matrix with true positives (TP) at 30, false negatives (FN) at 24, false positives (FP) at 12, and true negatives (TN) at 42. Using Equation (3), the recall was computed as 55.56%. A higher recall value meant missing fewer traffic near-misses. Using Equation (4), the precision was calculated as 71.43%. A higher precision value indicated fewer instances of mistakenly identifying safe driving scenarios as traffic near-misses. However, a recall lower than the precision indicated an improvement in detecting traffic near-misses because some traffic near-misses still remained unnoticed.

Table 1 - Confusion matrix

| Prediction / Ground truth | Near-miss | Safe driving |
|---|---|---|
| **Near-miss** | 30 (TP) | 24 (FN) |
| **Safe driving** | 12 (FP) | 42 (TN) |

The main evaluation metrics were adopted as shown in Equations (2) to (5) to comprehensively evaluate the model performance. Accuracy is the ratio of the traffic near-miss incidents that the model correctly predicts. Recall is the model's ability to correctly identify all positive samples (actual traffic near-misses). Precision is the ratio of those predicted as positive traffic near-misses by the model, which were actually positive. The F1 score considers the harmonic mean of precision and recall, providing a trade-off assessment of both performance aspects.

$$\text{Accuracy} = \frac{TP+TN}{TP+TN+FP+FN} \qquad (2)$$

$$\text{Recall} = \frac{TP}{TP+FN} \qquad (3)$$

$$\text{Precision} = \frac{TP}{TP+FP} \qquad (4)$$

$$F1 = 2 \times \frac{\text{Precision} \times \text{Recall}}{\text{Precision}+\text{Recall}} \qquad (5)$$

The method from [3] included a temporal encoding layer using CNNs for the video frame analysis, a grid-embedding layer for identifying critical image regions, and a multitask layer integrating these features to classify the traffic near-misses. A soft attention mechanism focused on the important data across these layers. Compared to [3], our SlowFast network showed significant advantages. Table 2 shows that compared with the video-only method [3], our model improved by 26.88 points in precision, 12.43 points in recall, and 19.11 points in the F1 score. Using only video as the input, these results demonstrate that our model enhanced the ability to efficiently identify traffic near-misses. While [3] used 4200 data samples, we utilized only 287 and achieved better results with considerably fewer data samples.

This improved performance was believed to result from the efficient spatiotemporal information processing capability of the SlowFast network. Its structure, which separated fast and slow processing, enabled the model to effectively capture the dynamic changes and the complex scene information, which was essential for the accurate identification of the traffic near-misses.

We conducted a visual analysis of the model using DeepGazeIIE [6], a deep learning model that mimics human gaze, and the Grad-CAM [5] technique to understand the connection between human visual perception processes and deep learning model decision-making. Grad-CAM helped understand which regions of an image the model focused on when making decisions. We also applied DeepGazeIIE [6] to explore the connection between the model results and human vision. DeepGazeIIE, an attention heatmap technique that simulates human gaze, was trained on real human eye-tracking data and created a saliency map in which areas assumed to be more focused had higher values. We compared the Grad-CAM results of the SlowFast network with the heatmap generated by DeepGazeIIE to examine how the human visual mechanism gets involved in the model decision-making process. Figures 5 and 6 show the representative examples of this comparison.

Table 2 - Results of the proposed method
[*1]From [3] paper. [*2]V: video. [*3]VSO: video, sensor, and objects. [*4]Compared to the NTT video only.

| Method | Precision (%) | Recall (%) | F1(%) | Accuracy (%) |
|---|---|---|---|---|
| ST-CNN[*1] (V[*2]) | 49.25 | 51.03 | 48.99 | 51.03 |
| SVM[*1] (VSO[*3]) | 57.64 | 58.25 | 57.64 | 58.25 |
| NTT[*1] (V[*2]) | 44.55 | 43.13 | 43.39 | - |
| NTT[*1] (VSO[*3]) | 65.75 | **65.79** | **65.68** | 65.79 |
| Ours (V[*2]) | **71.43** (26.88↑) [*4] | 55.56 (12.43↑) [*4] | 62.50 (19.11↑) [*4] | **66.67** |

Figure 5 presents the results of the two pathways (i.e., Slow and Fast), the heatmap from DeepGazeIIE for the cases classified as safe driving, and the original image. A notable difference was observed in the distribution of attention between DeepGazeIIE and the Slow and Fast channels. DeepGazeIIE tended to focus on the vehicles ahead, while the Slow and Fast channels concentrated more on the parallel vehicles.

Figure 6 presents the results of the Slow and Fast channels along with the heatmap from DeepGazeIIE for cases classified as near-miss incidents.

DeepGazeIIE tended to focus on the vanishing point of the road ahead and the notable surrounding areas (e.g., house). The Slow channel was more focused on parts where cars were stopped at T-junctions and the locations where near-miss incidents were likely to occur. The Fast channel was more focused on the notable surrounding areas. Figure 7 demonstrates that just seconds later, a car appeared at the T-junction, almost

Using SlowFast Networks for Near-Miss Incident Analysis in Dashcam Videos

causing a traffic near-miss incident.

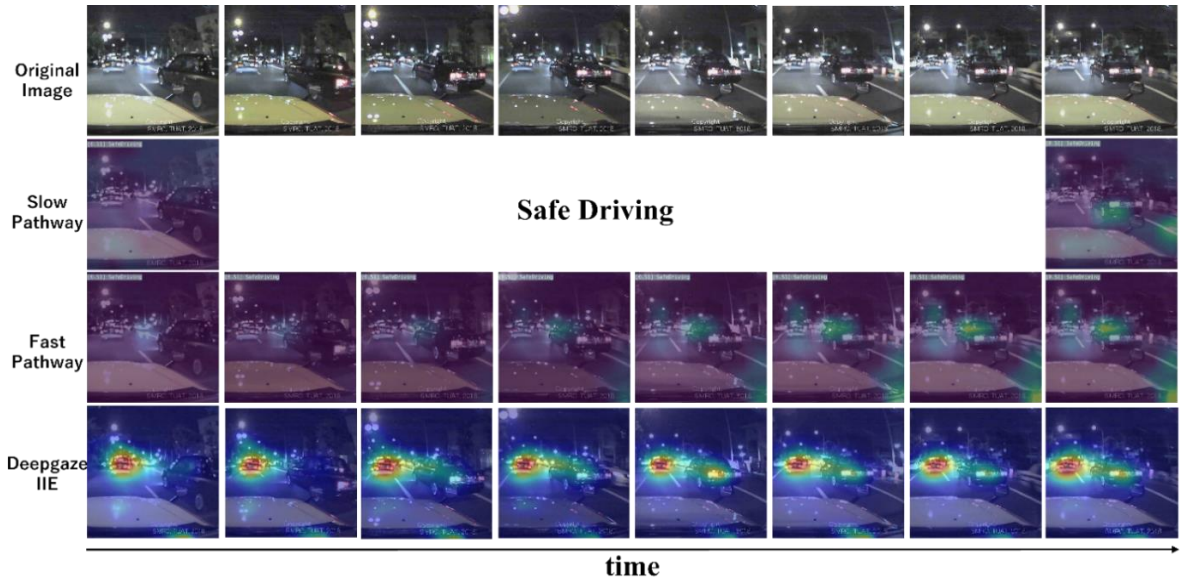

Figure 5 - Heatmap of safe driving

This difference may be caused by the two models designed with different considerations for different visual elements. Being trained in alignment with human gaze data, DeepGazeIIE tended to result in a bias toward the distant or central field of view objects, which can lead to missing attention on other parts and potentially cause near-miss incidents. In contrast, the Slow and Fast channels were sensitive to nearby or surrounding dynamic objects. Notably, despite being trained without human gaze data, the Fast channel produced results overlapping with the DeepGazeIIE heatmap, suggesting that the Fast channel effectively captured dynamic information and mimicked human visual functions.

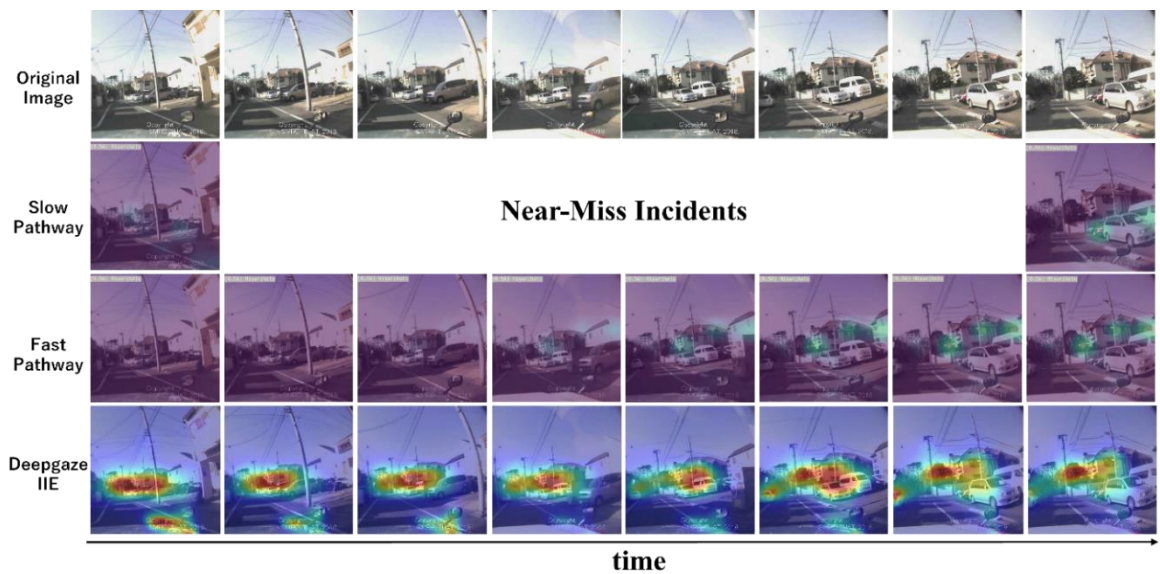

Figure 6 - Heatmap of the near-miss incident

Based on these findings, we suggest focusing on the forward attention of DeepGazeIIE and the peripheral dynamic detection of the Slow and Fast channels when predicting the near-miss incidents. This approach allowed the DeepGazeIIE model to predict where human drivers were likely to concentrate their gaze in different scenes, indicating which areas might attract a driver's attention. Combining these gaze data with the video frames processed by the SlowFast model provided additional contextual information for each frame, which might enable the model to detect potential near-miss incidents in areas where human drivers were not paying attention. Increasing interaction and data sharing between the models can also reduce their respective limitations and enhance the overall system safety and accuracy. In the end, this multi-model fusion approach aims to effectively reduce the near-miss incidents during driving, providing a more comprehensive safety measure for driving.

**Several seconds later, near-miss incidents occurred.**

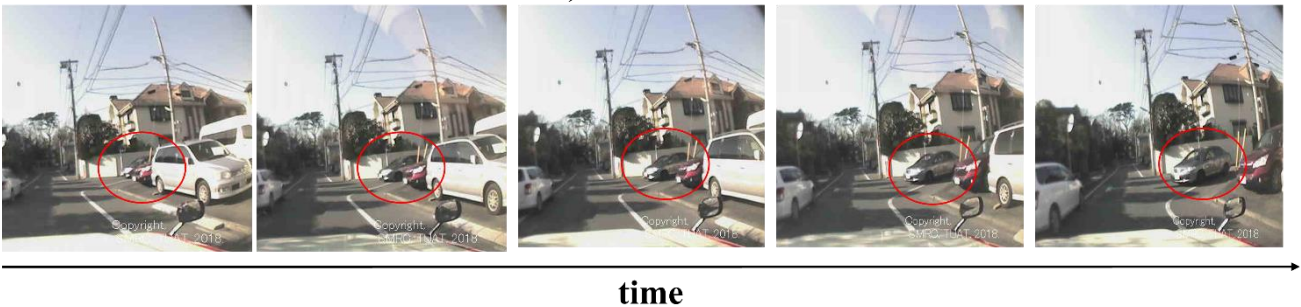

time

Figure 7 - Moment of a near-miss occurrence.

Although the model showed a good performance in the current experiment, several limitations and areas for improvement must be considered. First, we have the dataset issue. Only 287 data samples were used here, which was a very small dataset for deep learning models. Increasing the dataset may lead to better results. Second, we have the real-time performance issue. When implementing the model in vehicle systems, its real-time processing capability needs further optimization because current deep learning models still pose a significant burden to vehicle systems. Third, we have the explanation issue. While the model accuracy was good, the explanation of its decision-making process still needed improvement, which is crucial for actual deployment and user trust.

**Conclusion**

In this study, we conducted an in-depth analysis of the Near-Miss Incident dataset using the SlowFast network and proposed a new method for the near-miss incident classification. This method creatively utilized the biological principles of human visual information processing, especially the mechanisms by which the M and P cells process fast and slow visual information, thereby significantly enhancing the classification accuracy in the traffic near-miss incident videos. Compared to existing research [3], the SlowFast network adopted in this study demonstrated a higher accuracy in recognizing the traffic near-misses depending solely on the video input. The experimental results showed that after training with 287 data samples, the model achieved 66.67% accuracy on the test dataset, depicting an outstanding classification ability between near-miss incidents and safe driving. By combining Grad-CAM with the DeepGazeIIE heatmap, we also explored more thoroughly into the human visual mechanism in the model decision-making process. Future work can integrate the analytical results of DeepGazeIIE and SlowFast, focusing more effectively on areas that are often overlooked by humans and further improving the identification ratio of near-miss incidents.

This study achieved fine accuracy results, but had the following limitations:
1. Dataset increase: The current dataset is small. The size must be increased to improve the model generalization ability and accuracy.
2. Optimization of the real-time processing capabilities: The real-time processing capabilities must be optimized to improve the model performance in real-time systems, especially in vehicle systems.
3. Improvement of model explanation: Enhancing the transparency and the explanation of the model decision-making process is important in increasing the user's trust in the model.

Regarding research on human vision, further exploration and analysis will be considered using PredNet as used in [7][8][9], which mimics human illusions.

**Acknowledgment**

**References**

1. Tokyo University of Agriculture and Technology (TUAT). (2015). The copyright of the 'Near-Miss Database' has been fully transferred to our university. Retrieved from [link[*]] (In Japanese)

    link[*]: https://www.tuat.ac.jp/NEWS/2015/20150810_01.html
2. Raksincharoensak, P. (2013, October). Drive recorder database for accident/incident study and its potential for active safety development. In *FOT-NET Workshop*.
3. Yamamoto, S., Kurashima, T., & Toda, H. (2022). Classifying Near-Miss Traffic Incidents through Video, Sensor, and Object Features. *IEICE TRANSACTIONS on Information and Systems*, *105*(2), 377-386.
4. Feichtenhofer, C., Fan, H., Malik, J., & He, K. (2019). Slowfast networks for video recognition. In *Proceedings of the IEEE/CVF International Conference on Computer Vision* (pp. 6202-6211).
5. Selvaraju, R. R., Cogswell, M., Das, A., Vedantam, R., Parikh, D., & Batra, D. (2017). Grad-cam: Visual explanations from deep networks via gradient-based localization. In *Proceedings of the IEEE/CVF International Conference on Computer Vision* (pp. 618-626).
6. Linardos, A., Kümmerer, M., Press, O., & Bethge, M. (2021). DeepGaze IIE: Calibrated prediction in and out-of-domain for state-of-the-art saliency modeling. In *Proceedings of the IEEE/CVF International Conference on Computer Vision* (pp. 12919-12928).
7. Watanabe, E., Kitaoka, A., Sakamoto, K., Yasugi, M., & Tanaka, K. (2018). Illusory motion reproduced by deep neural networks trained for prediction. *Frontiers in psychology*, 345.
8. Emura, K., Kato, M., & Watanabe, E. (2022). Deep Neural Networks that Reproduce Human Vision Perception While Driving. *Transactions of the Society of Automotive Engineers of Japan. Nov2022*, Vol. 53 Issue 6, p1102-1107. (In Japanese)
9. Kato, M., Emura, K., & Watanabe, E. (2022). Analysis of Factors Causing Traffic Near-miss Events using

Deep Neural Networks Trained to Simulate Human Vision. *Transactions of the Society of Automotive Engineers of Japan. Nov2022*, Vol. 53 Issue 6, p1108-1113. (In Japanese)

10. Zeki, S. (2015). Area V5—a microcosm of the visual brain. Frontiers in integrative neuroscience, 9, 21.
11. Derrington, A. M., & Lennie, P. (1984). Spatial and temporal contrast sensitivities of neurones in lateral geniculate nucleus of macaque. The Journal of physiology, 357(1), 219-240.
12. Livingstone, M., & Hubel, D. (1988). Segregation of form, color, movement, and depth: anatomy, physiology, and perception. Science, 240(4853), 740-749.
13. National Police Agency (Japan). "Statistics about Road Traffic." Retrieved from [link[*]].
    link*: https://www.npa.go.jp/publications/statistics/koutsuu/toukeihyo_e.html
14. He, K., Zhang, X., Ren, S., & Sun, J. (2016). Deep residual learning for image recognition. In *Proceedings of the IEEE conference on computer vision and pattern recognition* (pp. 770-778).
15. Wang, X., Girshick, R., Gupta, A., & He, K. (2018). Non-local neural networks. In *Proceedings of the IEEE conference on computer vision and pattern recognition* (pp. 7794-7803).